\documentclass[journal]{IEEEtran}
\usepackage{amsmath,amsfonts}
\usepackage{algorithmic}
\usepackage{array}
\usepackage[caption=false,font=normalsize,labelfont=sf,textfont=sf]{subfig}
\usepackage{textcomp}
\usepackage{stfloats}
\usepackage{url}
\usepackage{verbatim}
\usepackage{graphicx}
\usepackage{cite}
\usepackage[colorlinks=true, linkcolor=blue]{hyperref}
\usepackage{cleveref}
\usepackage[table]{xcolor}
\usepackage{graphicx} 
\usepackage{wrapfig}
\usepackage{bm}
\usepackage{multirow}
\usepackage{algorithm}
\usepackage{algorithmic}
\usepackage{amsmath}       
\usepackage{booktabs} 
\usepackage{amssymb}
\definecolor{RevisionRed}{RGB}{0,0,0}

\newcommand{\revision}[1]{\textcolor{RevisionRed}{#1}}
\definecolor{AQERevisionRed}{RGB}{255,0,0}
\newif\ifshowaqerevisions
\ifdefined\AQEMarked
  \showaqerevisionstrue
\else
  \showaqerevisionsfalse
\fi
\newcommand{\aqerevision}[1]{%
  \ifshowaqerevisions
    \textcolor{AQERevisionRed}{#1}%
  \else
    #1%
  \fi
}

\newcommand{\pub}[1]{{\color{gray} #1}}
\def\BibTeX{{\rm B\kern-.05em{\sc i\kern-.025em b}\kern-.08em
    T\kern-.1667em\lower.7ex\hbox{E}\kern-.125emX}}
\usepackage{balance}
\begin{document}

\title{OMINI: Towards Consistent 3D Object Manipulation via Generative Texturing and Adaptive Lighting}

\author{Ruisi Zhao, Zechuan Zhang, Zongxin Yang, and Yi Yang,~\IEEEmembership{Fellow, IEEE}
\thanks{Ruisi Zhao, Zechuan Zhang, and Yi Yang are with the ReLER, CCAI, Zhejiang University, Hangzhou 310027, China (e-mail: zhaors00@zju.edu.cn; zechuan@zju.edu.cn; yangyics@zju.edu.cn).
\par Zongxin Yang is with the College of Artificial Intelligence, Zhejiang University, Hangzhou, China (e-mail: yangzongxin.ai@zju.edu.cn).

}}

\markboth{IEEE Transactions on Image Processing}%
{Zhao \MakeLowercase{\textit{et al.}}: OMINI: Towards Consistent 3D Object Manipulation}

\maketitle

\begin{abstract}
Enabling object manipulation based on a single static image, from geometric editing to animation, remains a significant challenge. Existing methods often struggle to jointly support these tasks, while also suffering from fidelity degradation in object appearance and scene illumination. In this work, we introduce \textit{OMINI}, a unified framework that addresses these challenges through two complementary strategies. First, OMINI lifts the 2D object into a 3D representation and adopts a \textit{Refine-before-Render} design. Both geometric editing and animation are then performed by manipulating and rendering the refined 3D asset, maintaining appearance consistency throughout the manipulation process. Second, we propose two dedicated modules for fidelity enhancement: \textit{TextuRefiner} refines rendered views of the 3D asset under appearance and geometric guidance, thereby generating fine-grained texture details; \textit{IllumiCorrector} corrects environmental illumination using HSL-based color correction and VLM-guided intensity optimization, enabling visually plausible renderings with shadows. Extensive experiments demonstrate that OMINI outperforms existing methods, producing visually consistent results. The GPU stages of OMINI are run on a single NVIDIA RTX 3090.
\end{abstract}

\begin{IEEEkeywords}
Diffusion models, image editing and animation.
\end{IEEEkeywords}
\section{Introduction} \label{sec:introduction}

\par \IEEEPARstart{O}{bject} manipulation
\cite{kholgade20143d, weng2019photo, yenphraphai2024image} plays an important
role in visual content creation, with applications in poster design, AR/VR,
and filmmaking. For example, it can present a product from different
viewpoints in an AR preview or animate a static subject in a film frame. The growing demand has driven the development of generative methods to make it easier and more efficient.

\begin{figure}[th!]
    \centering
    \includegraphics[width=\linewidth]{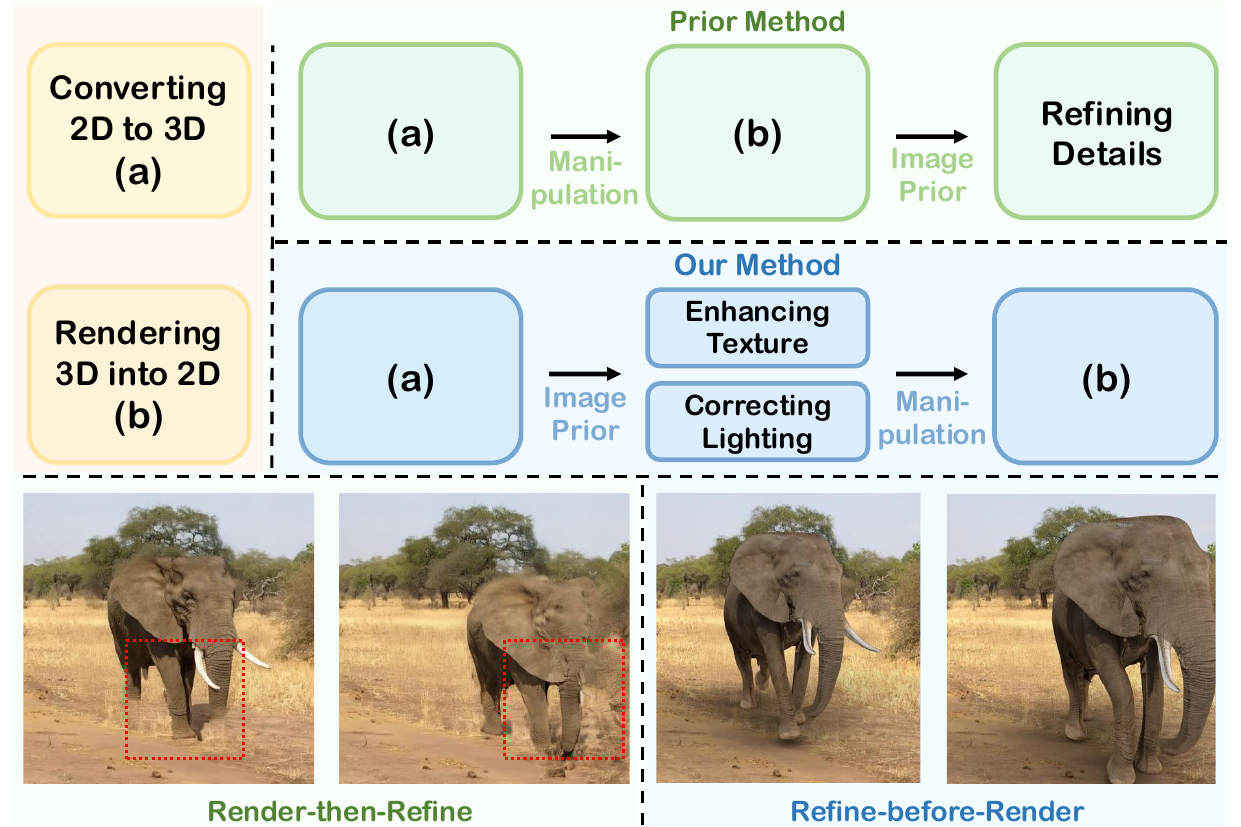}  
    \vspace{-0.6 cm}
    \caption{\textbf{Motivation of our method (\S\ref{sec:introduction}).}
Previous 3D-based object manipulation frameworks
\cite{yenphraphai2024image, hou2024training} adopt a
\textit{Render-then-Refine} design (top), applying 2D refinement to individual
rendered outputs. This design can introduce appearance inconsistencies across
rendered views and overwrite the geometry-aware shading produced by 3D
rendering, as highlighted by the
\textcolor[RGB]{204, 0, 0}{red} box. In contrast, OMINI adopts a
\textit{Refine-before-Render} design (bottom), refining the asset texture and
correcting environmental illumination before rendering. The refined asset is
then manipulated and rendered under the corrected illumination, maintaining
appearance consistency throughout the manipulation process and producing
visually plausible renderings with shadows.}
    
    \vspace{-0.6 cm}
    \label{fig:highlevel}
\end{figure}

\par Despite notable progress, prior methods still face two limitations in object manipulation:  1) \textit{\textbf{Difficulty in supporting both geometric editing and animation}}. Current image editing approaches \cite{yenphraphai2024image, michel2023object, hertz2022prompt, cao2023masactrl}, while effective for modifying object appearance, struggle to perform geometric edits and to maintain temporal consistency when extended to animation. Conversely, dynamic generation methods \cite{yang2024cogvideox, xing2023dynamicrafter}, while capable of producing plausible motion, offer limited control over specific visual edits.  
2) \textit{\textbf{Lack of fidelity in object appearance and scene illumination.}} Recent 2D techniques often distort object appearance during manipulation.
To address this issue, some methods
\cite{yenphraphai2024image, hou2024training} adopt 3D representations to
improve appearance consistency, but often sacrifice texture fidelity. Furthermore, generative methods largely overlook illumination. Rendering-based approaches instead model it explicitly using estimated environmental lighting~\cite{zhan2021emlight, phongthawee2024diffusionlight}. However, these estimates often fail to match
the original scene, resulting in unrealistic color tones.

\par To address these challenges, we propose two complementary strategies. \textbf{First}, to unify geometric editing and animation, our framework reconstructs the 2D object of interest as a 3D asset. This unified representation provides a common foundation for performing both tasks through operations in 3D space. \textbf{Second}, to enhance the fidelity of object appearance and scene illumination, we propose a decoupled refinement strategy: \textit{TextuRefiner} generates fine-grained texture details, while \textit{IllumiCorrector} corrects environmental illumination to produce visually plausible renderings with shadows. Together, these strategies produce high-fidelity results across tasks by rendering the manipulated 3D asset with its refined texture and corrected environmental illumination.

\par Building on the above strategies, we detail the design of our framework, \textbf{OMINI} (\underline{\textbf{O}}bject \textbf{\underline{M}}anipulation in a s\underline{\textbf{IN}}gle \underline{\textbf{I}}mage). As illustrated in Fig.~\ref{fig:highlevel}, OMINI is directly motivated by the limitations of prior 3D-based methods \cite{yenphraphai2024image, hou2024training}. These methods employ a \textit{Render-then-Refine} design, applying 2D refinement to individual rendered outputs. This design tends to introduce cross-view appearance inconsistencies, cause flickering during manipulation, and overwrite the geometry-aware shading produced by 3D rendering. In contrast, OMINI adopts the ``\textit{\textbf{Refine-before-Render}}'' design, in which the asset texture is enhanced before rendering. Both geometric editing and animation are then performed by manipulating and rendering the refined 3D asset, maintaining appearance consistency throughout the manipulation process. Furthermore, the subsequent 3D rendering stage naturally enables the integration
of environmental illumination, producing visually plausible renderings with shadows.

\par Additionally, we introduce \textit{\textbf{TextuRefiner}} to enhance texture fidelity and \textit{\textbf{IllumiCorrector}} to correct environmental illumination. Prior texturing methods \cite{li2023sweetdreamer, cao2023texfusion, richardson2023texture} face a dilemma: text-based approaches struggle to preserve identity without image conditioning, while image-driven approaches often suffer from fine-grained detail loss. To address this, \revision{TextuRefiner refines rendered views of the 3D asset under
appearance and geometric guidance. It adapts
SDXL~\cite{podell2023sdxl} with an object-specific DreamBooth
LoRA~\cite{hu2021lora, ruiz2023dreambooth} to retain appearance details,
while depth maps rendered from the 3D asset provide geometric conditioning.
The refined views are then used to optimize the texture map through
differentiable rasterization~\cite{laine2020modular}.} To address the limitations of existing illumination estimators \cite{zhan2021emlight, phongthawee2024diffusionlight}, IllumiCorrector employs a decoupled correction strategy. It first performs HSL-based color correction on the estimated environmental illumination, using the object's appearance as a reference. For intensity optimization, a Vision-Language Model (VLM) agent \cite{team2023gemini, achiam2023gpt, wu2026platform} evaluates object exposure and shadow visibility in the rendered result and iteratively adjusts the lighting intensity.

\par To validate the effectiveness of our framework, we evaluate OMINI on five representative manipulation tasks (\S\ref{sec:setup}), covering both geometric editing and animation. We also compare OMINI with a range of baselines, including editing
approaches (e.g., Image Sculpting~\cite{yenphraphai2024image} and
StableFlow~\cite{avrahami2025stable}) and animation approaches, including
general image-to-video models (e.g., CogVideoX~\cite{yang2024cogvideox} and
Wan 2.1~\cite{wan2025wan}) and the trajectory-controlled method
DragAnything~\cite{wu2024draganything}. Quantitative comparisons and a user study~(\S\ref{sec:quantitative}) show that OMINI achieves strong overall performance across tasks. Detailed ablation studies~(\S\ref{sec:ablation}) further validate the effectiveness of TextuRefiner and IllumiCorrector. In summary, our main contributions are as follows:

\noindent\begin{itemize}
    \item We propose OMINI (\S\ref{sec:overview}), a unified framework built around the \textit{Refine-before-Render} design. Both geometric editing and animation are performed by manipulating and rendering the refined 3D asset, maintaining appearance consistency throughout the manipulation process.
    
    \item We introduce TextuRefiner (\S\ref{sec:texture}), which refines rendered views of the 3D asset under appearance and geometric guidance. The refined views are used to optimize the texture map through differentiable rasterization, generating fine-grained texture details lost during initial reconstruction.
    
    \item We design IllumiCorrector (\S\ref{sec:light}), an illumination correction module that combines HSL-based color correction with VLM-guided intensity optimization to produce visually plausible renderings with shadows.
    
    \item Extensive quantitative and qualitative experiments (\S\ref{sec:experiments}) demonstrate that OMINI outperforms existing baselines on both geometric editing and animation tasks.
\end{itemize}

\section{Related Work}

\subsection{Image Editing}

\par Diffusion models \cite{ho2020denoising, peebles2023scalable, rombach2021highresolution, podell2023sdxl} have been widely applied to image editing
\cite{brooks2023instructpix2pix, hertz2022prompt, mokady2023null,
tumanyan2023plug, zhang2026enabling}. Existing methods can be broadly grouped into training-free and fine-tuning based approaches. Among training-free methods, MasaCtrl \cite{cao2023masactrl} and Plug-and-Play (PnP) \cite{tumanyan2023plug} improve structural consistency by reusing source-image attention or intermediate features during inference. RF-Solver \cite{wang2024taming} improves inversion accuracy by reducing
ODE-solving errors, whereas StableFlow \cite{avrahami2025stable} identifies
critical MM-DiT blocks and selectively injects features into them. In contrast, fine-tuning based methods update the diffusion U-Net~\cite{ronneberger2015u} using prepared editing data. InstructPix2Pix
\cite{brooks2023instructpix2pix} and MagicBrush
\cite{zhang2023magicbrush}, for example, learn from image pairs and editing
instructions to perform instruction-guided edits. Beyond 2D editing, several
methods incorporate 3D priors to improve geometric control
\cite{kholgade20143d, michel2023object}. Image Sculpting
\cite{yenphraphai2024image} uses 3D depth cues to guide generation and
subsequently refines the output in 2D.

\subsection{Image Animation}

\par With the rise of video diffusion models~\cite{ho2022video}, numerous
methods use diverse conditioning signals to guide video generation~\cite{wang2024videocomposer, videofusion2023,
wu2024customcrafter}. Some methods \cite{shi2024motion, mahapatra2023text} combine estimated optical flow with a pre-trained text-to-image diffusion model~\cite{rombach2021highresolution}
to generate videos. Other works focus on controllable editing through motion conditions, including object movement \cite{chen2024motion} and camera movement \cite{yang2024direct}. StoryDiffusion \cite{zhou2024storydiffusion} combines consistent
self-attention with a semantic motion predictor to maintain subject consistency and generate smooth transitions. More recently, open-source models such as CogVideoX \cite{yang2024cogvideox}, HunyuanVideo \cite{kong2024hunyuanvideo}, and Wan 2.1 \cite{wan2025wan} support image-based video generation, leading a trend towards more powerful, general-purpose solutions. However, current video generation methods still have significant room for improvement in maintaining both object appearance consistency and motion fluidity.

\subsection{Illumination Harmonization}

\par Illumination harmonization is a crucial vision task that ensures objects are naturally integrated into new backgrounds, and many studies \cite{zhan2021emlight, wang2022stylelight, wang2022neural, phongthawee2024diffusionlight} focus on addressing this challenge. IntrinsicCompos \cite{careaga2023intrinsic} adjusts the foreground albedo to align with the background and estimates environmental lighting to refine the shading process. DiffRefMap \cite{enyo2024diffusion} introduces a technique that utilizes 3D object information to generate high-quality reflectance maps, enabling recovery of detailed lighting information. DiPIR \cite{liang2024photorealistic} estimates environmental illumination from a background image but requires optimization for each scene and inserted object. ICLight \cite{zhang2025scaling} achieves impressive results in generating arbitrary lighting effects; however, it is not fully suitable for our task as it often alters the background, making it less applicable for video tasks. Due to the plug-and-play nature of our framework, environment maps estimated by methods such as DiPIR could also be used as
inputs to IllumiCorrector.

\section{Method}
\label{sec:method}
\begin{figure*}[th!]
  \centering
  \includegraphics[width=\linewidth]{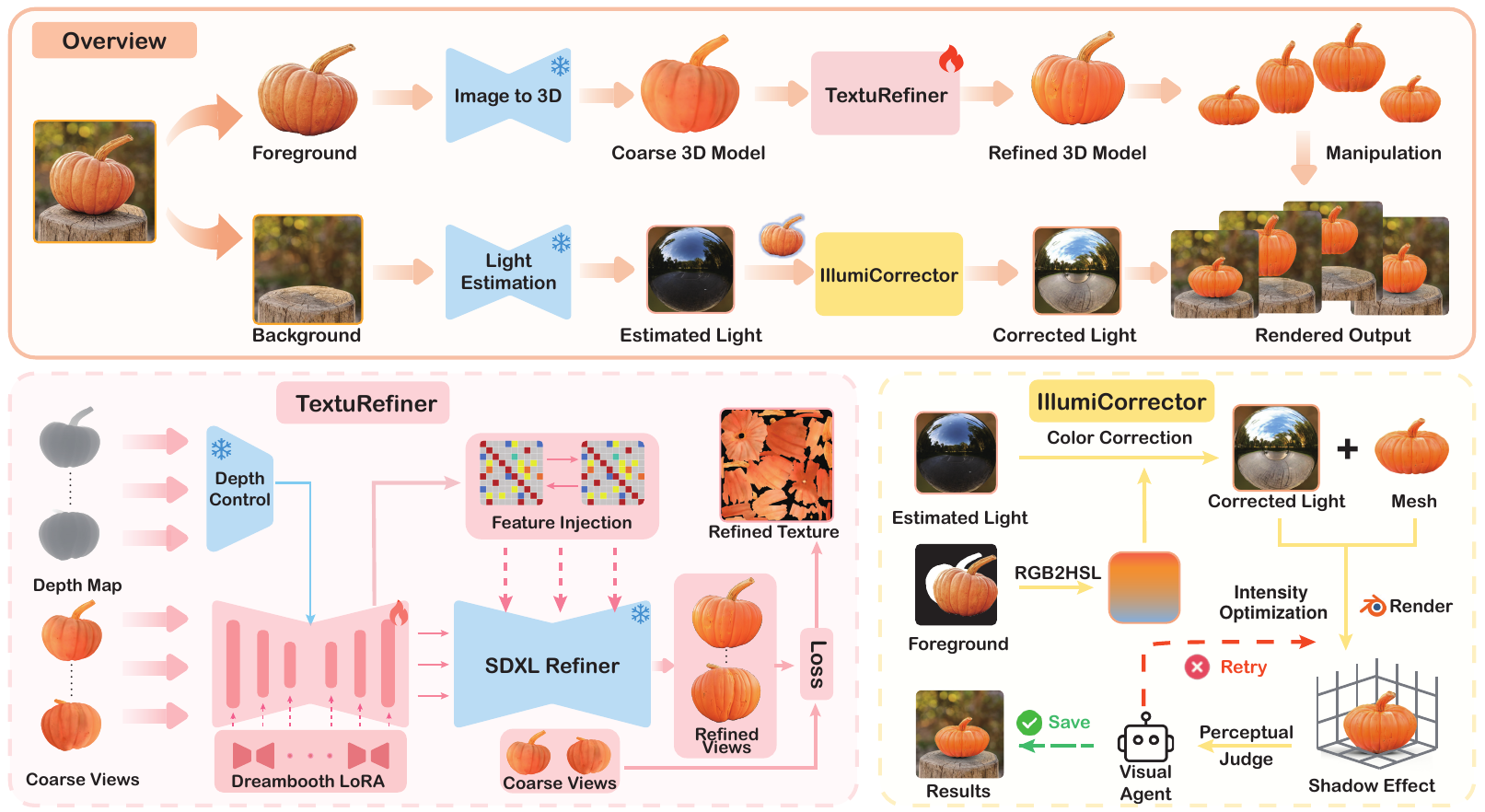}
  \vspace{-0.7 cm}

  \caption{\textbf{Overview of the OMINI framework (\S\ref{sec:method}).} Our framework follows a \textit{Refine-before-Render} design that unifies geometric editing and animation through a shared 3D representation. The process begins by lifting the 2D object into a coarse 3D asset, followed by two key modules: (1) \textit{TextuRefiner}, which refines rendered views of the 3D asset under appearance and geometric guidance to generate fine-grained texture details; and (2) \textit{IllumiCorrector}, which corrects the estimated environmental illumination through object-aware HSL-based color correction and VLM-guided intensity optimization, enabling visually plausible renderings with shadows. Finally, the refined 3D asset is rendered under the corrected environmental illumination using Blender~\cite{blender}, producing geometry-aware and visually coherent manipulation results.}

  \vspace{-0.5 cm}
  \label{fig:pipeline}
\end{figure*}

\par In this section, we first provide brief preliminaries on diffusion models in Section~\ref{sec:preliminary}. We then present an overview of our \textbf{OMINI} framework in Section~\ref{sec:overview}, with the pipeline illustrated in Fig.~\ref{fig:pipeline}. Finally, we describe \textbf{\textit{TextuRefiner}} and \textbf{\textit{IllumiCorrector}} in Section~\ref{sec:texture} and Section~\ref{sec:light}, respectively.

\subsection{Preliminary}
\label{sec:preliminary}

\par\noindent\textbf{Diffusion models} \cite{ho2020denoising,song2020denoising,zhao2023unipc} model a data distribution $p_{\mathrm{data}}$ by reversing a progressive noising process. The forward diffusion is modeled as a discrete-time stochastic process $\{ \bm{x}_t \}_{t=0}^T$ where $\bm{x}_0 \sim p_{\mathrm{data}}$, and $\bm{x}_t \mid \bm{x}_{t-1} \sim \mathcal{N}(\sqrt{\alpha_t / \alpha_{t-1}} \bm{x}_{t-1}, (1 - \alpha_t / \alpha_{t-1})\bm{I})$. The decreasing scalar function $\alpha_t$, with constraints that $\alpha_0 = 1$ and $\alpha_T \approx 0$, controls the noise level over time. It can be shown that:
\begin{equation}
\label{eq:add_noise}
\bm{x}_t
=
\sqrt{\alpha_t}\bm{x}_0
+
\sqrt{1-\alpha_t}\bm{\epsilon},
\bm{\epsilon}\sim\mathcal{N}(\bm{0},\bm{I}).
\end{equation}
A diffusion model $\epsilon_{\phi}$ is trained to predict the noise
$\bm{\epsilon}$ from the noisy sample $\bm{x}_t$. Its simplified training
objective is defined as:
\begin{equation}
\label{eq:diffusion_objective}
\mathcal{L}
=
\mathbb{E}_{\bm{x}_0,t,\bm{\epsilon}}
\left[
\left\|
\epsilon_{\phi}(\bm{x}_t,t,\bm{C})
-
\bm{\epsilon}
\right\|_2^2
\right],
\end{equation}
where $\bm{C}$ denotes conditions such as text or images.

\subsection{Overview of OMINI}
\label{sec:overview}

\par\noindent\textbf{From 2D Object to Manipulable 3D Asset.}
As shown in Fig.~\ref{fig:pipeline}, OMINI begins by segmenting the
object of interest from an input image using SAM~\cite{kirillov2023segment}.
The segmented foreground is then processed by an off-the-shelf image-to-3D
model~\cite{liu2023zero, wu2024unique3d, zhang2024clay,
zhang2023globalcorrelated, Zhang_2024_CVPR, xiang2025structured} to obtain a
coarse 3D asset, which provides a shared 3D representation for both geometric
editing and animation. In contrast to prior methods \cite{yenphraphai2024image, hou2024training} that post-process individual renderings, OMINI adopts the \textit{Refine-before-Render} design: \textit{TextuRefiner} generates fine-grained details in the asset's texture map before rendering. The refined asset is then used throughout subsequent manipulation and rendering, preserving appearance consistency. \revision{For rigid transformations and composition, the asset's rotation, position, and scale can be directly adjusted in 3D space. \aqerevision{Deformation and animation are manually configured and adjusted,
with optional support from simulation or rigged motion sequences generated using existing tools~\cite{song2025magicarticulate,mixamo}.} Consequently, each new
manipulation reuses the same refined asset and requires only the
corresponding 3D operations and rendering, resulting in a low
computational cost.}


\noindent
\textbf{Geometry-Aware Rendering with Illumination Correction.} OMINI automatically calibrates the camera to align the reconstructed object with the input image. The camera parameters are first estimated through a coarse search comparing the projected foreground mask with
the SAM mask, then refined using Powell's method and retained for subsequent rendering. With the spatially aligned 3D scene established, environmental illumination can be explicitly incorporated during rendering. \textit{IllumiCorrector} therefore corrects the estimated environmental illumination to better match the scene. After manipulation, the refined asset is rendered under the corrected illumination and composited with the original background, producing visually plausible renderings with shadows.

\subsection{Cross-view Generative Texture Refinement} \label{sec:texture}

\par To generate fine-grained texture details lost during initial reconstruction, we introduce \textbf{\textit{TextuRefiner}}. It refines rendered views of the 3D asset under appearance and geometric guidance. These refined views then serve as targets for optimizing the texture map through differentiable rasterization~\cite{laine2020modular}. We describe its design details below.

\par\noindent\textbf{View Refinement via Object-Specific Diffusion.} \revision{To provide object-specific appearance guidance, \textit{TextuRefiner} adapts SDXL~\cite{podell2023sdxl} with an object-specific DreamBooth LoRA~\cite{hu2021lora, ruiz2023dreambooth} using a single
foreground image segmented from the input image.} For multi-view refinement, we render the coarse 3D asset from multiple viewpoints and invert each rendered view into the latent noise space using DDIM Inversion~\cite{song2020denoising}. During denoising, the adapted diffusion model refines each view while preserving the identity and appearance of the
input object across viewpoints.

\par\noindent\textbf{Geometric Guidance via Depth Control.}
To provide geometric guidance during view refinement, we render a corresponding depth map for each coarse view of the 3D asset using the same camera pose. Each depth map
is used to condition a \revision{frozen Depth ControlNet~\cite{zhang2023adding}} during both DDIM
Inversion~\cite{song2020denoising} and diffusion sampling, maintaining geometric consistency in the refined views. To further retain fine-grained appearance details, we adopt a feature-injection strategy inspired by PnP~\cite{tumanyan2023plug}. Feature maps from the residual blocks and self-attention layers of the U-Net are stored during inversion and injected into the corresponding layers during denoising.

\par\noindent\textbf{Texture Optimization.}
\revision{The refined views then serve as targets for optimizing the texture map through differentiable rasterization. Let $u$ denote the texture map to be optimized. For each camera pose $p$, the differentiable rasterizer renders the 3D asset using the current texture map $u$, producing $I_{\mathrm{render}}^p$. We optimize the texture map using the following objective:
\begin{equation}
\label{eq:uv_optimization}
\mathcal{L}_{\mathrm{total}}
=
\sum_p
\left\|
V^p \odot
\left(
I_{\mathrm{fine}}^p-I_{\mathrm{render}}^p
\right)
\right\|_2^2,
\end{equation}
where $I_{\mathrm{fine}}^p$ denotes the refined target view at camera pose $p$, and $V^p$ denotes the visibility mask of the object from the corresponding viewpoint. The view-specific reconstruction losses are jointly accumulated across camera poses to update the texture map, allowing it to capture complementary fine-grained details from multiple views.}

\subsection{Adaptive Illumination Correction}
\label{sec:light}

\par We introduce \textbf{\textit{IllumiCorrector}}, an illumination correction module designed to improve the visual compatibility of the estimated environmental illumination. It combines HSL-based color correction
with VLM-guided intensity optimization to produce visually plausible renderings with shadows.

\par\noindent\textbf{Color Correction.}
Following previous illumination estimation approaches
\cite{zhan2021emlight, phongthawee2024diffusionlight}, we obtain an initial
illumination map $Light_{\mathrm{env}}$ from the background. In the rendering pipeline, $Light_{\mathrm{env}}$ is decomposed into a color component
$c_{\mathrm{env}}$, which determines the illumination color, and an intensity component $i_{\mathrm{env}}$, which controls its overall strength. However, the initial estimate may not visually match the scene, leading to desaturated
colors or inconsistent brightness in the rendered object.

\par To mitigate these artifacts and improve visual coherence, we introduce an \textit{\textbf{object-aware HSL-based color correction}} strategy. It selects informative pixels from the object and uses their saturation and
lightness statistics to correct the color component $c_{\mathrm{env}}$ while leaving its hue unchanged. We describe its design details below.

\par First, we segment the object using SAM~\cite{kirillov2023segment} to obtain an object mask $M$ and convert the segmented region to the HSL (Hue, Saturation, Lightness) color space. \revision{We determine the thresholds $T_S$ and $T_L$ from the saturation and lightness percentiles within the object region, respectively. The saturation threshold $T_S$ is computed from a low percentile to remove near-white specular highlights and gray regions with insufficient color information. The lightness threshold $T_L$ is computed from a high percentile to retain bright object regions and reduce the influence of shadows.} The resulting target region is defined as:
\[
M_k = \{x \in M \mid S(x) \ge T_S \text{ and } L(x) \ge T_L \}.
\]

\par We next average the saturation and lightness values within the selected region $M_k$ to obtain the object-derived targets $S_{\mathrm{target}}$ and $L_{\mathrm{target}}$, respectively:
\begin{equation}
S_{\mathrm{target}}
=
\frac{1}{|M_k|}
\sum_{x \in M_k} S(x),
\quad
L_{\mathrm{target}}
=
\frac{1}{|M_k|}
\sum_{x \in M_k} L(x),
\end{equation}
where $|M_k|$ denotes the number of pixels in $M_k$. We also compute the mean saturation $\bar{S}_{\mathrm{env}}$ and mean lightness $\bar{L}_{\mathrm{env}}$ over the initial illumination map $Light_{\mathrm{env}}$.

\par Based on these statistics, we define
$\widetilde{S}_{\mathrm{env}}$ and $\widetilde{L}_{\mathrm{env}}$ as the
desired saturation and lightness after correction. Each value is
obtained by interpolating between the object-derived target
and the mean statistic of the initial illumination map:
\begin{equation}
\widetilde{S}_{\mathrm{env}}
=
\lambda_s S_{\mathrm{target}}
+
(1-\lambda_s)\bar{S}_{\mathrm{env}},
\end{equation}
\begin{equation}
\widetilde{L}_{\mathrm{env}}
=
\lambda_l L_{\mathrm{target}}
+
(1-\lambda_l)\bar{L}_{\mathrm{env}}.
\end{equation}
Here, $\lambda_s$ and $\lambda_l$ control the strengths of saturation and
lightness alignment, respectively.

\par We then convert these target levels into scaling factors:
\begin{equation}
\eta_s
=
\frac{\widetilde{S}_{\mathrm{env}}}
{\bar{S}_{\mathrm{env}}+\epsilon},
\qquad
\eta_l
=
\frac{\widetilde{L}_{\mathrm{env}}}
{\bar{L}_{\mathrm{env}}+\epsilon},
\end{equation}
where $\epsilon$ is a small constant for numerical stability. The same
factors are applied to the saturation and lightness channels of every pixel
$x$ in the illumination map:
\begin{equation}
S_{\mathrm{env}}'(x)
=
\eta_s S_{\mathrm{env}}(x),
\qquad
L_{\mathrm{env}}'(x)
=
\eta_l L_{\mathrm{env}}(x).
\end{equation}

The corrected saturation and lightness values are clipped to their valid ranges, while the hue remains unchanged
($H_{\mathrm{env}}'(x)=H_{\mathrm{env}}(x)$). The corrected channels are then recombined to obtain the color-corrected illumination map used for subsequent rendering.

\begin{figure*}[th!]
  \centering
  \includegraphics[width=\linewidth]{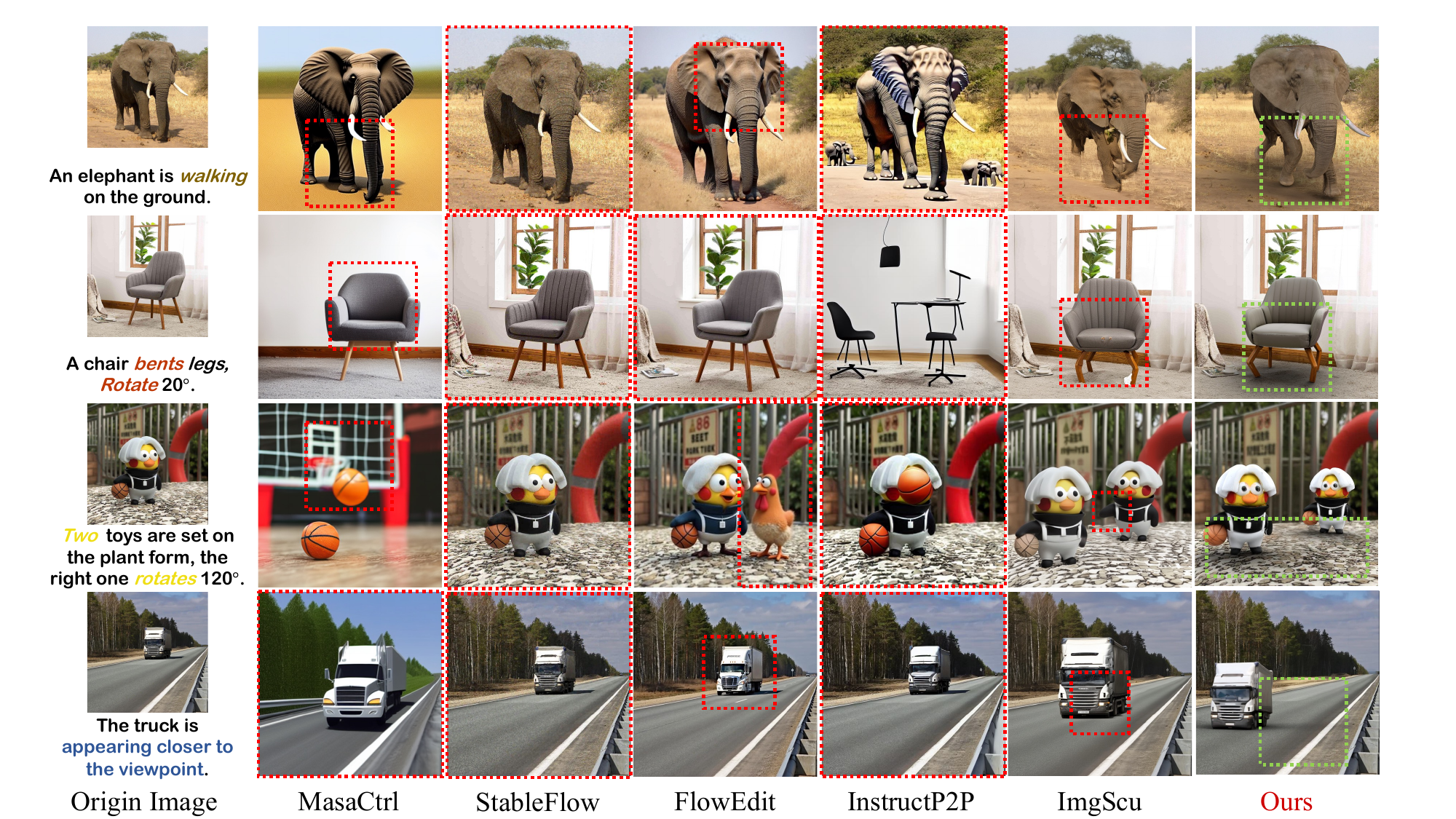}
  \vspace{-0.7 cm}
  \caption{\textbf{Qualitative comparison with image editing methods} (\S\ref{sec:qualitative}). \revision{InstructP2P denotes InstructPix2Pix.} Image Sculpting (ImgScu) exhibits structural distortions and missing content, such as distorted elephant legs and a missing basketball, while StableFlow often leaves the input nearly unchanged. Failure cases and unrealistic shadows are marked with \textcolor[RGB]{234, 51, 35}{red} boxes, whereas visually plausible shadows produced by OMINI are marked with \textcolor[RGB]{159, 206, 99}{green} boxes.}
  \vspace{-0.5 cm}
  \label{fig:edit_compare}
\end{figure*}

\par\noindent\textbf{VLM-Guided Intensity Optimization.}
After applying HSL-based color correction to the estimated environmental illumination, we further optimize its global intensity to improve the visual compatibility between the rendered object and the surrounding scene. Let
$i_{\mathrm{env}}$ denote its initial intensity. We parameterize the optimized intensity as:
\begin{equation}
i_{\mathrm{final}}=\gamma i_{\mathrm{env}},
\end{equation}
where $\gamma$ is a global scaling factor to be optimized.

\par We determine $\gamma$ through an iterative VLM-guided evaluation process. At each iteration, the system renders a result using the current $\gamma$. The VLM agent~\cite{team2023gemini} evaluates object exposure and
shadow visibility in the rendered result and updates $\gamma$ accordingly. Applying the optimized intensity $i_{\mathrm{final}}$ to the color-corrected illumination map produces the final illumination map $Light_{\mathrm{final}}$.

\par During rendering, a shadow-catcher plane is placed beneath the 3D asset. It captures the object's contact shadows under $Light_{\mathrm{final}}$ without appearing in the rendered image. The rendered object and its shadows are then composited with the original background to produce the final result.

\section{Experiments}
\label{sec:experiments}
\subsection{Experimental Setup}

\par\noindent\textbf{Evaluation Data and Tasks.}
\label{sec:setup}
\revision{Our evaluation set comprises 20 cases, including examples from the public PhysGen3D evaluation set~\cite{chen2025physgen3d}
and additional cases collected by us.} The evaluation covers five tasks: four geometric editing tasks---\textit{Deformation},
\textit{Composition}, \textit{Relocation}, and \textit{Rotation}---and
animation. Results for geometric editing and animation are reported
separately.

\begin{figure*}[th!]
  \centering
  \includegraphics[width=\linewidth]{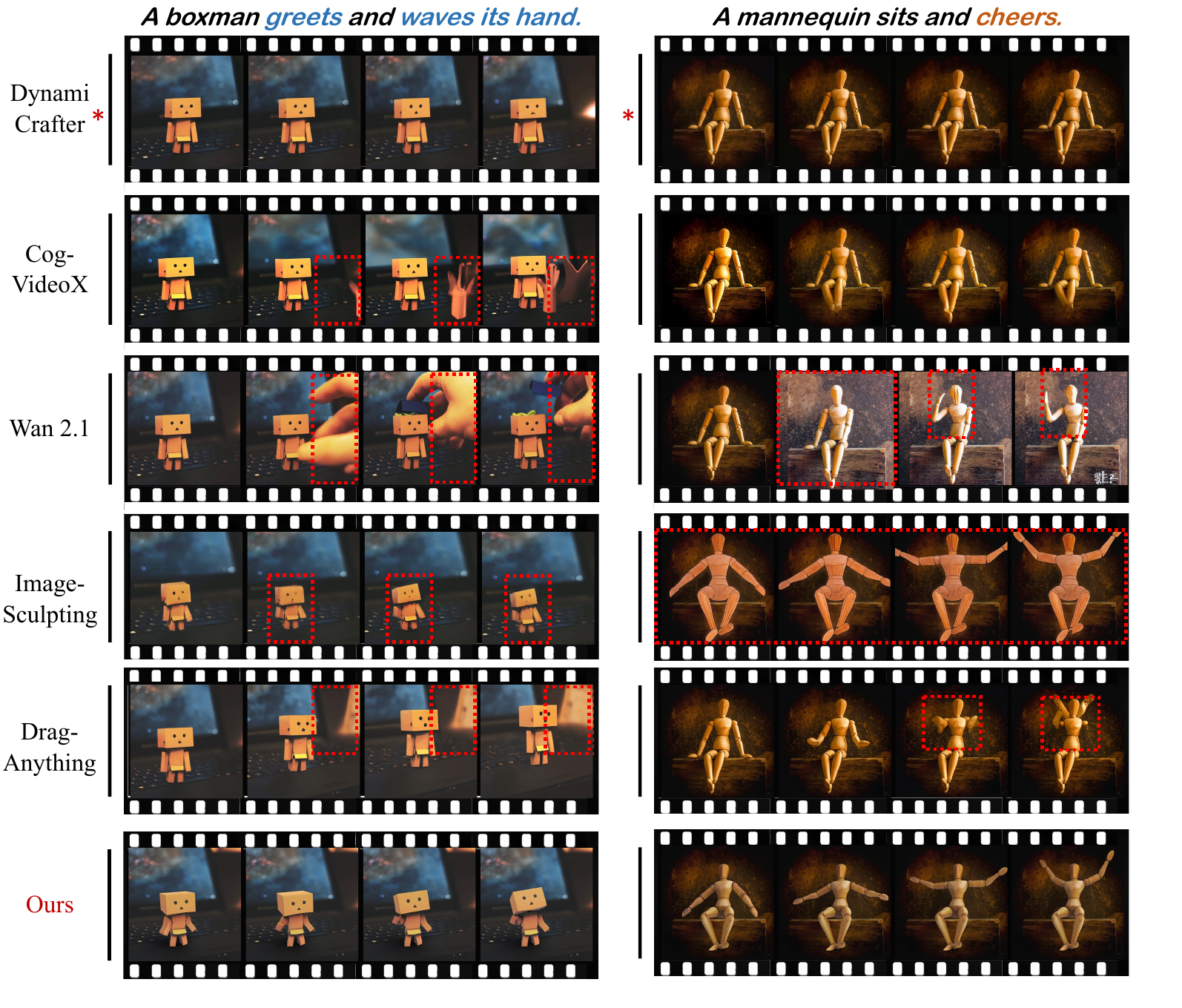}
  \vspace{-1.0 cm}
  \caption{\textbf{Qualitative comparison with image animation methods
(\S\ref{sec:qualitative}).} A \textcolor[RGB]{204, 0, 0}{red asterisk}
marks cases in which the object remains nearly stationary despite the
requested motion. Distortions are highlighted by
\textcolor[RGB]{204, 0, 0}{red} rectangles. DragAnything still struggles
with fine-grained motion control, sometimes causing global camera motion or
object distortion. We further zoom in on the wooden mannequin result from
Image Sculpting to show its noticeable appearance variations.}
  \vspace{-0.5 cm}
  \label{fig:video compare}
\end{figure*}

\par\noindent\textbf{Implementation Details.}
We first use SAM~\cite{kirillov2023segment} to segment the object of interest.
We primarily adopt Rodin~\cite{hyper3d_rodin} to reconstruct the initial mesh
and texture map, while OMINI is also compatible with other image-to-3D
reconstruction frameworks~\cite{liu2023zero, wu2024unique3d,
zhao2025hunyuan3d, xiang2025structured}. 
For TextuRefiner, we adapt SDXL~\cite{podell2023sdxl} with a
concept-specific DreamBooth LoRA~\cite{ruiz2023dreambooth, hu2021lora}.
\revision{The LoRA is trained \aqerevision{with rank 4} using only
the segmented foreground image at a resolution of $1024\times1024$
for 3,000 steps, with a learning rate of $6\times10^{-5}$.}
\aqerevision{DDIM inversion uses 100 steps, and refinement uses
100 denoising steps with a classifier-free guidance scale of 5.0.}
\revision{For multi-view refinement, we use six principal views:
four horizontal views at azimuth angles of
$\{0^\circ,90^\circ,180^\circ,270^\circ\}$, together with a top view
and a bottom view.}
\aqerevision{The refinement images are generated at
$1024\times1024$, and the UV texture map is resampled to
$4096\times4096$ before optimization.}
For IllumiCorrector, we use DiffusionLight~\cite{phongthawee2024diffusionlight} to estimate the initial environmental illumination and set $\lambda_s=0.5$ and $\lambda_l=0.3$ for
HSL-based correction. The intensity scaling factor $\gamma$ is initialized to 1.0 and iteratively adjusted by a VLM agent based on Gemini 2.5 Pro~\cite{team2023gemini}. \aqerevision{The GPU stages of OMINI are run on a single NVIDIA RTX 3090.} 

\par\noindent\aqerevision{For automatic camera calibration, a coarse
search is performed over azimuth angles at $15^\circ$ intervals,
elevation angles of $\{-15^\circ,0^\circ,15^\circ,30^\circ,45^\circ\}$,
and focal lengths of $\{35,50,70\}$ mm. For each candidate, camera
distance and lens shifts are adjusted to align the projected object
with the SAM mask. Based on mask alignment, three candidate camera
configurations are selected for separate Powell-based refinement
to improve overlap between the projected foreground mask and the
SAM mask, with the best-aligned configuration retained for subsequent
rendering. For background preparation, the original target is removed
and the resulting region is inpainted once~\cite{team2023gemini,
achiam2023gpt}. The prepared background is reused for subsequent edits,
with the rendered foreground composited onto it using transparent
RGBA output. For final rendering, static results for qualitative
presentation are rendered at $1024\times1024$, while the static
quantitative evaluation uses $512\times512$ renderings. For animation,
our evaluation uses clips of approximately 5 seconds at 16 FPS,
matching the Wan \cite{wan2025wan} baseline setting.}


\par\noindent\textbf{Baselines.}
For geometric editing, we compare OMINI with five representative methods:
MasaCtrl~\cite{cao2023masactrl}, StableFlow~\cite{avrahami2025stable},
FlowEdit~\cite{kulikov2025flowedit}, InstructPix2Pix \revision{(InstructP2P)}~\cite{brooks2023instructpix2pix}, and Image Sculpting (ImgScu)~\cite{yenphraphai2024image}. Among them, Image Sculpting incorporates explicit 3D control and therefore provides a relevant comparison with our 3D-based manipulation framework. For animation, we compare with three general image-to-video models: DynamiCrafter~\cite{xing2023dynamicrafter}, CogVideoX~\cite{yang2024cogvideox}, and Wan 2.1~\cite{wan2025wan}, as well as two controlled baselines: Image Sculpting, a 3D-controlled \textit{Render-then-Refine} method, and \revision{DragAnything~\cite{wu2024draganything}}, a trajectory-controlled method. All methods receive the same source image, while the desired manipulation is specified through each method's control interface.

\par\noindent\textbf{Evaluation Metrics.}
For geometric editing, we report CLIP Directional Similarity
(CLIP-Dir)~\cite{radford2021learning,gal2022stylegan}, which measures how
well the visual change from the input image to the edited result follows the
requested manipulation in the CLIP embedding space. Higher values indicate
better alignment. \revision{For animation, we use RAFT~\cite{teed2020raft} to align adjacent
frames according to their estimated motion. LPIPS~\cite{zhang2018unreasonable}
is then computed between the aligned frames, reducing the influence of
intended object motion and measuring the remaining perceptual appearance
changes. We also compute pixel-wise warping error to evaluate low-level
temporal consistency. Lower values indicate better temporal consistency.}

\par\noindent\textbf{GPT-5.6-Based Evaluation.}
Since feature-based metrics do not fully capture high-level perceptual
quality, we additionally conduct a method-blind evaluation using
\revision{GPT-5.6}~\aqerevision{\cite{openai_gpt56}} with a fixed prompt and scoring
rubric. For each geometric editing result, the evaluator receives the input image, the manipulation instruction, and the generated image without any method identity. It assigns integer scores from 1 to 5 along two dimensions. \textit{Image Alignment} assesses the preservation of object identity and appearance, as well as visual coherence with the scene, while \textit{Text Alignment} measures how well the result follows the requested manipulation. For each animation result, the evaluator receives the input image, the motion instruction, and five temporally ordered frames sampled from the animation. The evaluator additionally assesses temporal \textit{Consistency}, considering the stability of object identity, texture, shape, and illumination across frames. \revision{All automated and GPT-5.6 scores are averaged over the complete evaluation set.}

\begin{table*}[th!]
\centering
\setlength{\tabcolsep}{8pt}
\renewcommand{\arraystretch}{1.1}

\caption{\textbf{Quantitative comparison on geometric editing
({\fontshape{n}\selectfont\S}\ref{sec:quantitative}).}
We report CLIP-Dir as a feature-based metric, together with GPT-5.6 and
user-study scores on a scale of 1--5. $\uparrow$ and $\downarrow$ indicate
that higher and lower values are better, respectively. The best and
second-best results are shown in \textbf{bold} and
\underline{underlined}, respectively.}

\vspace{-0.3 cm}
\small
\begin{tabular}{@{}l|c|cc|cc@{}}
\toprule
& \textbf{Automated}
& \multicolumn{2}{c|}{\textbf{GPT-5.6 Evaluation}}
& \multicolumn{2}{c}{\textbf{User Study}} \\
\cmidrule(lr){2-2}
\cmidrule(lr){3-4}
\cmidrule(lr){5-6}

\textbf{Method}
& CLIP-Dir $\uparrow$
& Image Align $\uparrow$
& Text Align $\uparrow$
& Image Align $\uparrow$
& Text Align $\uparrow$ \\
\midrule

\rowcolor[HTML]{F9FFF9}
MasaCtrl \cite{cao2023masactrl}~\pub{ICCV23}
& 0.0250
& 2.71
& 2.56
& 2.65
& 2.11 \\

\rowcolor[HTML]{F9FFF9}
StableFlow \cite{avrahami2025stable}~\pub{CVPR25}
& 0.0150
& \textbf{3.74}
& 2.09
& \textbf{4.57}
& 1.26 \\

\rowcolor[HTML]{F9FFF9}
FlowEdit \cite{kulikov2025flowedit}~\pub{ICCV25}
& 0.0073
& 3.18
& 2.14
& 3.73
& 2.59 \\

\rowcolor[HTML]{F9FFF9}
InstructPix2Pix \cite{brooks2023instructpix2pix}~\pub{CVPR23}
& 0.0116
& 3.19
& 2.21
& 1.79
& 1.86 \\

\rowcolor[HTML]{F9FFF9}
Image Sculpting \cite{yenphraphai2024image}~\pub{CVPR24}
& \underline{0.0284}
& 3.20
& \underline{3.30}
& 3.68
& \underline{3.37} \\

\midrule

\rowcolor[HTML]{F9FBFF}
\textbf{Ours}
& \textbf{0.0336}
& \underline{3.36}
& \textbf{4.30}
& \underline{4.01}
& \textbf{4.27} \\

\bottomrule
\end{tabular}

\vspace{-0.4 cm}
\label{tab:quantitative evaluation image}
\end{table*}

\par\noindent\textbf{User Study.}
\revision{We invite ten Ph.D. candidates with research backgrounds in
generative models. Each participant evaluated ten cases for each of the four
geometric editing tasks and ten animation cases, with the compared methods presented in randomized order. Participants rated \textit{Image Alignment} and \textit{Text Alignment} for geometric editing,
and additionally rated temporal \textit{Consistency} for animation, using a
five-point scale. \aqerevision{Participants considered shadow plausibility and
consistency with the surrounding scene as part of their assessment
of overall visual coherence.} The final scores for each method are averaged across
participants and evaluation cases.}

\subsection{Quantitative Comparison}
\label{sec:quantitative}

\par\noindent\textbf{Geometric Editing Comparison.}
As shown in Tab.~\ref{tab:quantitative evaluation image}, OMINI achieves the highest CLIP-Dir score, followed by Image Sculpting (ImgScu), demonstrating stronger alignment with the requested geometric edits. In the GPT-5.6 evaluation and user study, OMINI also achieves the highest Text Alignment scores while maintaining competitive Image Alignment. StableFlow obtains high Image Alignment but substantially lower Text Alignment, indicating that it often leaves the input largely unchanged rather than performing the requested edit. Overall, OMINI better balances performing the
requested geometric edits and preserving a visually coherent object
appearance.

\par\noindent\textbf{Animation Comparison.}
As shown in Tab.~\ref{tab:quantitative evaluation video}, OMINI achieves the lowest flow-aligned LPIPS and warping error, while Image Sculpting ranks second on both metrics. In the GPT-5.6 evaluation, OMINI achieves the highest Consistency score
and a tied second-highest Text Alignment score; in the user study, it achieves the highest scores on both criteria. DynamiCrafter obtains high Image Alignment but considerably lower Text Alignment, suggesting that it often produces nearly static results rather than performing the requested animation. Although Image Sculpting performs well on the automated metrics, it receives the lowest Consistency score in the user study because of visible flickering and cross-frame appearance misalignment. Overall, OMINI preserves object appearance while producing temporally coherent animations.

\subsection{Qualitative Comparison}
\label{sec:qualitative}
\par \par To complement the quantitative results, we compare OMINI with baselines on multiple editing tasks. As shown in Fig.~\ref{fig:edit_compare}, existing methods often struggle to perform the requested edits while preserving object identity and structure. The red boxes highlight structural degradation, incomplete edits, and implausible shadows. StableFlow leaves the input largely unchanged apart from minor noise artifacts, thus failing to perform the requested transformations. Image Sculpting (ImgScu) performs some requested edits but introduces artifacts associated with its \textit{Render-then-Refine} design, including distorted elephant legs and a missing basketball. It also produces
implausible or missing shadows in several cases. In contrast, OMINI performs the requested geometric edits while preserving object identity and structure. As highlighted by the green boxes, OMINI also produces visually plausible shadows that naturally ground the manipulated object in the scene.

\begin{table*}[th!]
\centering
\setlength{\tabcolsep}{2pt}
\renewcommand{\arraystretch}{1.1}

\caption{\textbf{Quantitative comparison on image animation
({\fontshape{n}\selectfont\S}\ref{sec:quantitative}).}
We report flow-aligned LPIPS and pixel-wise warping error for objective
temporal-consistency evaluation. GPT-5.6 and user-study scores are reported
on a scale of 1--5. $\uparrow$ and $\downarrow$ indicate that higher and
lower values are better, respectively. The best and second-best results are
shown in \textbf{bold} and \underline{underlined}, respectively.}

\vspace{-0.2 cm}

\small
\begin{tabular}{@{}l|cc|ccc|ccc@{}}
\toprule
& \multicolumn{2}{c|}{\textbf{Automated}}
& \multicolumn{3}{c|}{\textbf{GPT-5.6 Evaluation}}
& \multicolumn{3}{c}{\textbf{User Study}} \\
\cmidrule(lr){2-3}
\cmidrule(lr){4-6}
\cmidrule(lr){7-9}

\textbf{Method}
& LPIPS $\downarrow$
& Warp Error $\downarrow$
& Image Align$\uparrow$
& Text Align$\uparrow$
& Consistency $\uparrow$
& Image Align$\uparrow$
& Text Align$\uparrow$
& Consistency $\uparrow$ \\
\midrule

\rowcolor[HTML]{F9FFF9}
DynamiCrafter \cite{xing2023dynamicrafter}~\pub{ECCV24}
& 0.0767
& 0.0137
& 4.21
& 2.14
& 4.50
& 3.79
& 1.63
& 2.79 \\

\rowcolor[HTML]{F9FFF9}
CogVideoX \cite{yang2024cogvideox}~\pub{ICLR25}
& 0.0862
& 0.0151
& 3.86
& \underline{2.64}
& 4.00
& 3.83
& 2.41
& 3.43 \\

\rowcolor[HTML]{F9FFF9}
Wan 2.1 \cite{wan2025wan}~\pub{arXiv25}
& 0.0704
& 0.0155
& 3.64
& \textbf{3.07}
& 3.86
& \textbf{3.96}
& 3.24
& 3.91 \\

\rowcolor[HTML]{F9FFF9}
DragAnything \cite{wu2024draganything}~\pub{ECCV24}
& 0.0949
& 0.0133
& 4.00
& 2.50
& \underline{4.79}
& 3.87
& \underline{3.41}
& \underline{3.98} \\

\rowcolor[HTML]{F9FFF9}
Image Sculpting \cite{yenphraphai2024image}~\pub{CVPR24}
& \underline{0.0396}
& \underline{0.0070}
& \underline{4.43}
& 2.21
& 4.64
& 2.94
& 2.56
& 2.15 \\

\midrule

\rowcolor[HTML]{F9FBFF}
\textbf{Ours}
& \textbf{0.0350}
& \textbf{0.0054}
& \textbf{4.64}
& \underline{2.64}
& \textbf{4.86}
& \underline{3.93}
& \textbf{3.57}
& \textbf{4.18} \\

\bottomrule
\end{tabular}

\vspace{-0.4 cm}
\label{tab:quantitative evaluation video}
\end{table*}

\begin{table}[t]
\centering
\setlength{\tabcolsep}{5pt}
\renewcommand{\arraystretch}{1.1}

\caption{\textbf{Quantitative comparison of texture-refinement methods.} CLIP-Image measures reference-view appearance similarity, while DINO measures foreground identity preservation. Higher values are better for both metrics ($\uparrow$).}
\vspace{-0.3 cm}
\small
\begin{tabular}{@{}l|cc@{}}
\toprule
\textbf{Method}
& CLIP-Image $\uparrow$
& DINO $\uparrow$ \\
\midrule

\rowcolor[HTML]{F9FFF9}
Paint3D~\cite{zeng2024paint3d}
& 0.9051
& 0.7123 \\

\rowcolor[HTML]{F9FFF9}
Hunyuan3D-Paint~\cite{zhao2025hunyuan3d}
& 0.9316
& 0.7301 \\

\midrule

\rowcolor[HTML]{F9FBFF}
\textbf{TextuRefiner (Ours)}
& \textbf{0.9471}
& \textbf{0.7823} \\

\bottomrule
\end{tabular}

\vspace{-0.6 cm}
\label{tab:texture_refinement_comparison}
\end{table}

\par \par For animation, we compare OMINI with several representative baselines, as shown in Fig.~\ref{fig:video compare}. Although these methods broadly preserve object identity, they often struggle to follow the requested motion. DynamiCrafter frequently leaves the object nearly static. CogVideoX and Wan 2.1 sometimes introduce unrelated elements, such as additional hands. \revision{DragAnything struggles with fine-grained motion control, sometimes producing global frame motion or object distortion.} Image Sculpting follows the requested motion more clearly, but its frame-by-frame refinement introduces noticeable texture variations and geometric misalignment across frames. In contrast, OMINI
applies the requested motion to a shared refined 3D asset, maintaining consistent texture and geometry across frames and producing coherent, visually plausible animations.

\subsection{Ablation Study}
\label{sec:ablation}
\par \par We conduct a series of ablation experiments to evaluate the core components of OMINI and validate their design choices. The analysis is organized around two modules. For \textbf{TextuRefiner}, we examine (a) \textbf{Overall Effect}, comparing the full module with a no-refinement baseline and alternative texture-refinement methods; and (b) \textbf{Component Analysis}, examining concept-specific DreamBooth LoRA, DDIM inversion, Depth ControlNet, and feature injection. For \textbf{IllumiCorrector}, we examine (a) \textbf{Overall Effect}, \aqerevision{comparing correction configurations and existing illumination methods}; (b) \textbf{Parameter Sensitivity}, analyzing the percentile levels used to compute the color-correction thresholds and the interpolation weights; and (c) \textbf{Optimization Robustness}, testing different initial values of $\gamma$ and analyzing the resulting optimization trajectories.

\begin{figure*}[ht!]
  \centering
  \includegraphics[width=\linewidth]{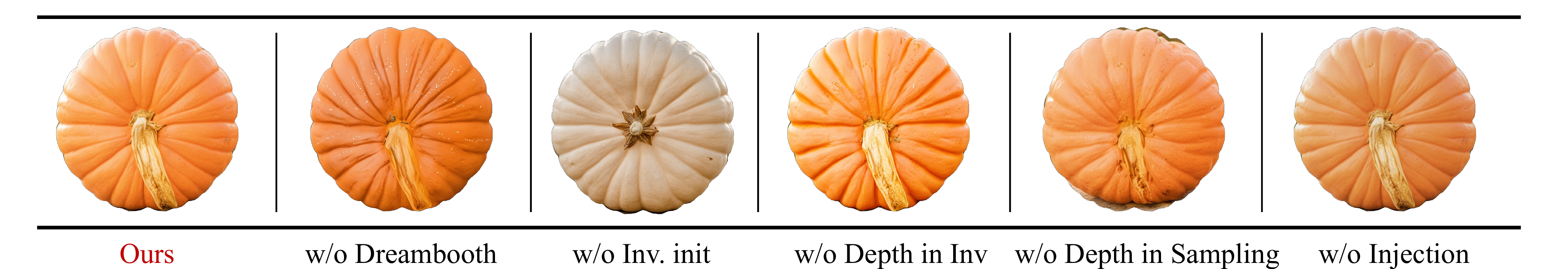}
  \vspace{-0.6 cm}
  
 \caption{\textbf{Component analysis of TextuRefiner (\S\ref{sec:ablation}).}
Results are visualized from a novel viewpoint. Compared with the full
TextuRefiner (leftmost), removing object-specific DreamBooth LoRA
(\textit{w/o DreamBooth}), \aqerevision{inversion-based initialization (\textit{w/o Inv. init})}, depth control during inversion
or sampling, or feature injection introduces identity drift, structural
artifacts, or color degradation.}
  
  \vspace{-0.4 cm}
  \label{fig:technique}
\end{figure*}
\begin{figure*}[th!]
\centering
\includegraphics[width=\linewidth]{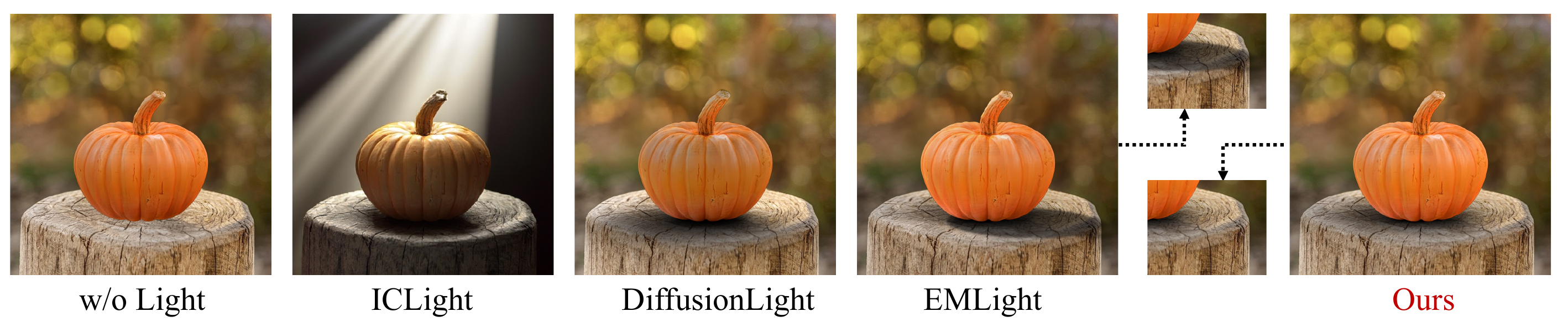}
\vspace{-0.7 cm}
\caption{\textbf{Qualitative comparison of IllumiCorrector with illumination estimation methods (\S\ref{sec:ablation}).}
EMLight produces noticeable overexposure, whereas DiffusionLight produces overly diffuse shadows. In contrast, IllumiCorrector produces object illumination that better matches the scene, together with clearer, geometry-aware contact shadows, as highlighted in the zoomed-in comparison.}
\vspace{-0.6 cm}
\label{fig:ablation_light}
\end{figure*}
\begin{figure}[t!]
\centering
\includegraphics[width=\linewidth]{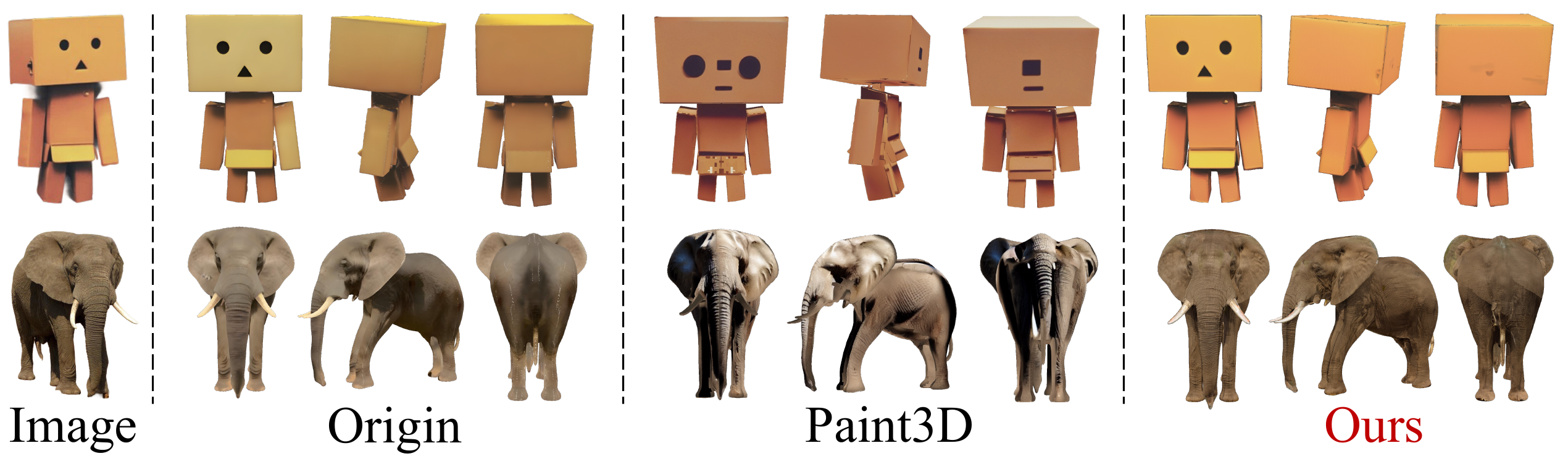}
\vspace{-0.5 cm}

\caption{\textbf{Qualitative comparison of texture-refinement methods} (\S\ref{sec:ablation}). \aqerevision{Origin} denotes the coarse texture obtained directly from image-to-3D reconstruction and serves as the no-refinement baseline. Compared with \aqerevision{Origin} and Paint3D, TextuRefiner recovers finer texture details while maintaining a more coherent appearance across viewpoints.}

\vspace{-0.7 cm}
\label{fig:ablation_texture}
\end{figure}
\begin{figure}[t!]
\centering
\includegraphics[width=0.9\linewidth]
{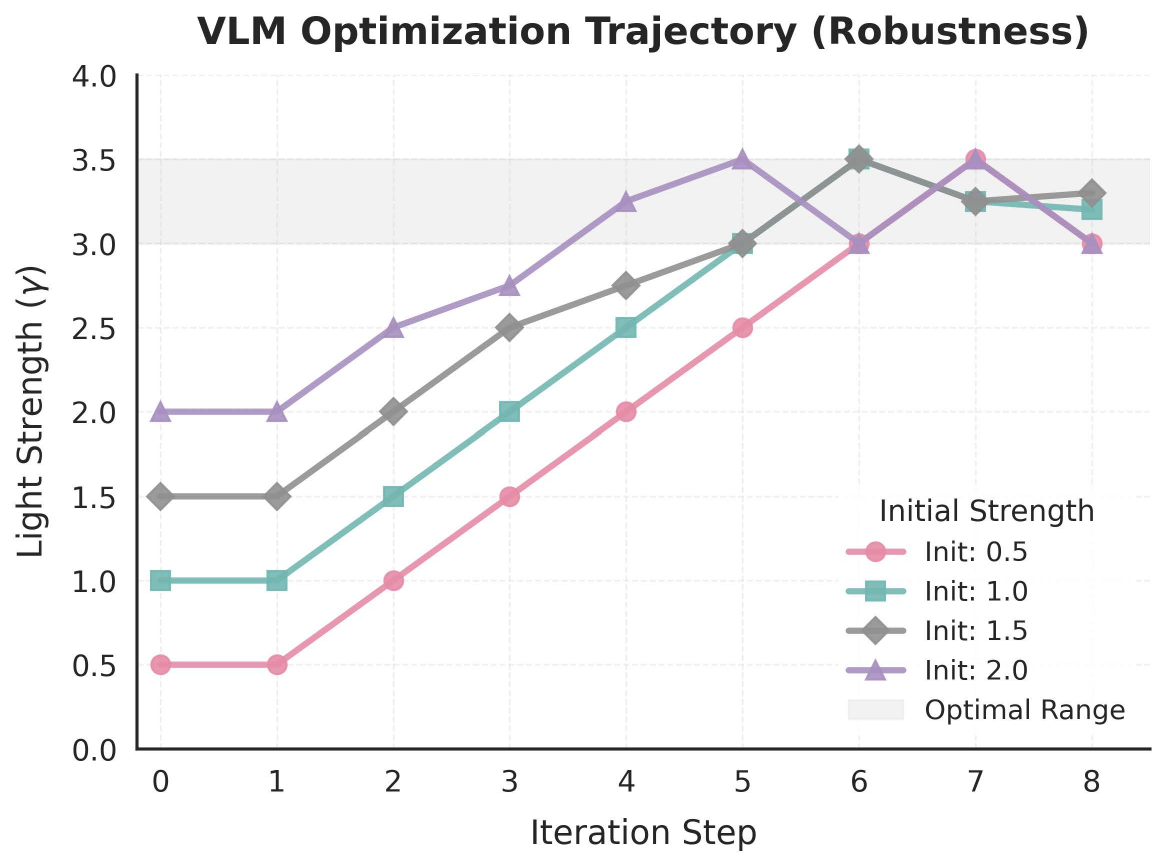}
\vspace{-0.4 cm}
\caption{\textbf{Convergence of the intensity scaling factor $\gamma$
(\S\ref{sec:ablation}).}
Starting from four different initial values, the VLM-guided optimization
drives $\gamma$ toward a similar stable range, indicated by the gray band,
within approximately six iterations. This result demonstrates the
robustness of the optimization to different initializations.}

\vspace{-0.6 cm}
\label{fig:ablation_vlm}
\end{figure}

\par\noindent\textbf{Overall Effect of TextuRefiner.}
We first qualitatively compare TextuRefiner with a no-refinement
reference and Paint3D across three representative viewpoints:
front, side, and back. As shown in Fig.~\ref{fig:ablation_texture},
the coarse texture produced by Rodin~\cite{hyper3d_rodin} is
over-smoothed and lacks texture details.
Paint3D~\cite{zeng2024paint3d} recovers additional details but may
introduce noise and appearance variations across viewpoints.
In contrast, TextuRefiner recovers finer texture details while
maintaining a more coherent object appearance across viewpoints.
\revision{We further compare TextuRefiner with Paint3D and
Hunyuan3D-Paint~\cite{zhao2025hunyuan3d} using the same 3D geometry
and rendering setup.}
\aqerevision{While these two methods rely on general-purpose
pretrained models for texture generation, TextuRefiner uses
per-object LoRA adaptation to tailor refinement to the input
object and preserve its fine-grained details.}
\revision{As reported in
Tab.~\ref{tab:texture_refinement_comparison}, TextuRefiner achieves
the highest CLIP-Image and DINO scores, indicating that its
reference-view renderings more closely match the input appearance
and better preserve foreground identity.}

\par\noindent\textbf{Component Analysis of TextuRefiner.}
\label{ablation_tech}
To examine the role of each design element in TextuRefiner, we visualize the ablation results from a novel viewpoint in Fig.~\ref{fig:technique}. First, removing concept-specific DreamBooth LoRA (\textit{w/o DreamBooth}) results in identity drift and produces an appearance inconsistent with the input object. \aqerevision{Second, initializing refinement from random noise
instead of the DDIM-inverted latent (\textit{w/o Inv. init}) leads
to severe appearance errors, producing a nearly gray pumpkin.} Furthermore, Depth ControlNet provides geometric guidance. Removing it during inversion (\textit{w/o Depth in Inv}) causes a loss of fine geometric details, as evidenced by the smoothed pumpkin grooves, while removing it during sampling
(\textit{w/o Depth in Sampling}) introduces structural artifacts and contour misalignment. Finally, removing feature injection
(\textit{w/o Injection}) reduces color saturation, causing the result to deviate from the input appearance. Together, these components improve appearance fidelity and geometric consistency.

\par\noindent\textbf{Effect of IllumiCorrector.}
\revision{\aqerevision{We compare three configurations in Tab.~\ref{tab:ablation_illumicorrector}: DiffusionLight alone, intensity optimization without HSL correction, and the complete IllumiCorrector.} For evaluation, we use CLIP-Image to assess the appearance similarity between the input object and its aligned reference-view rendering, and Foreground Color Error (FCE) to measure the CIEDE2000 color difference between the corresponding object regions. Higher CLIP-Image and lower FCE indicate better
appearance similarity and color reproduction, respectively. \aqerevision{Intensity optimization reduces FCE from 21.78 to 12.82, while the complete method further reduces it to 7.73 and achieves the highest CLIP-Image score. Comparing intensity optimization alone with the complete method supports the benefit of incorporating HSL-based color correction.}}

\par Fig.~\ref{fig:ablation_light} provides a qualitative comparison with existing illumination estimation methods. EMLight produces noticeable overexposure, whereas DiffusionLight produces overly diffuse shadows. In contrast, IllumiCorrector produces object illumination that better matches the scene.


\setcounter{topnumber}{1}
\makeatletter
\global\@topnum\@ne
\makeatother
\begin{figure*}[!tb]
  \centering
  \includegraphics[width=\linewidth]{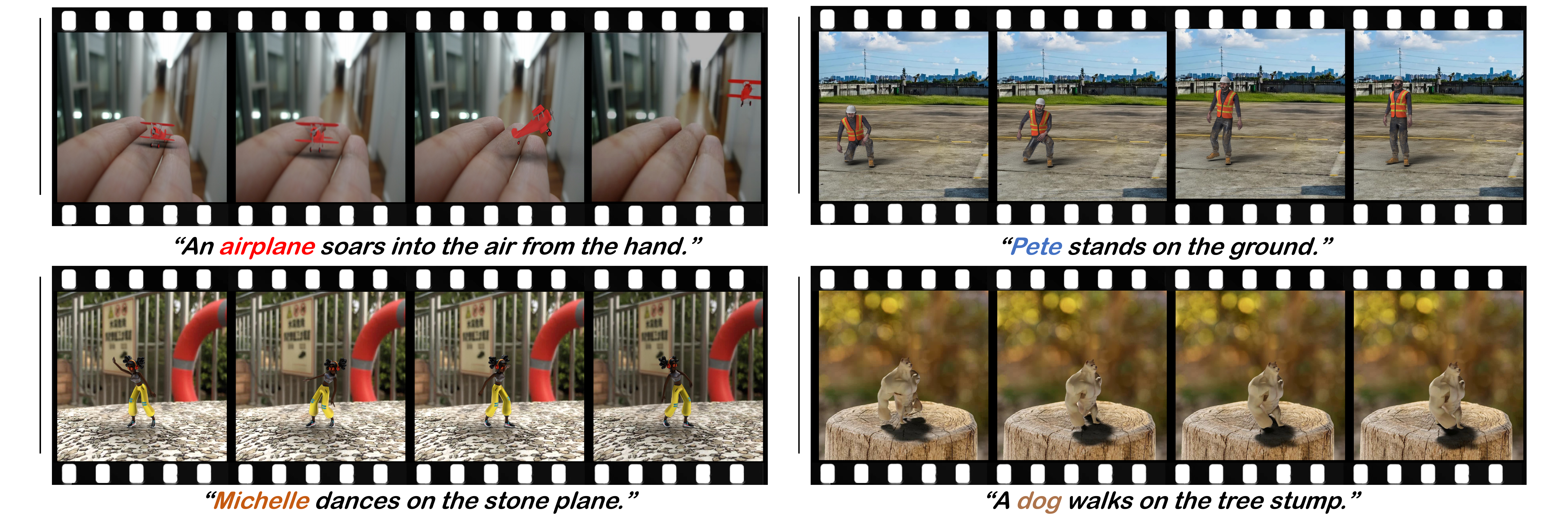}
  \vspace{-0.8 cm}
\caption{\textbf{Potential applications of OMINI
(\S\ref{sec:application}).} OMINI renders manipulated 3D assets under corrected environmental illumination and composites them with real-world backgrounds, enabling high-fidelity object insertion.}
  \label{fig:application}
  \vspace{-0.5 cm}
\end{figure*}
\begin{figure}[t]
    \centering
    \includegraphics[width=\columnwidth]{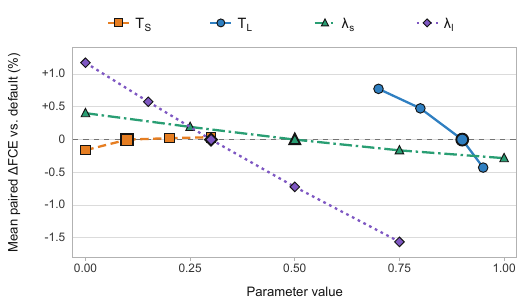}
    \vspace{-0.8 cm}
    
    \caption{\textbf{Parameter sensitivity of IllumiCorrector
(\S\ref{sec:ablation}).} Each curve varies one setting while keeping the others fixed at their default values. \revision{For $T_S$ and $T_L$, the horizontal axis denotes the percentile level used to compute the corresponding threshold.} For each setting, the percentage change in FCE relative to the default is computed for each case and averaged over cases. Enlarged markers indicate the default settings, while negative values indicate lower
FCE than the default.}

    \label{fig:parameter_sensitivity}
    \vspace{-0.5 cm}
    
\end{figure}


\begin{figure}[t]
\centering
\includegraphics[width=\linewidth]{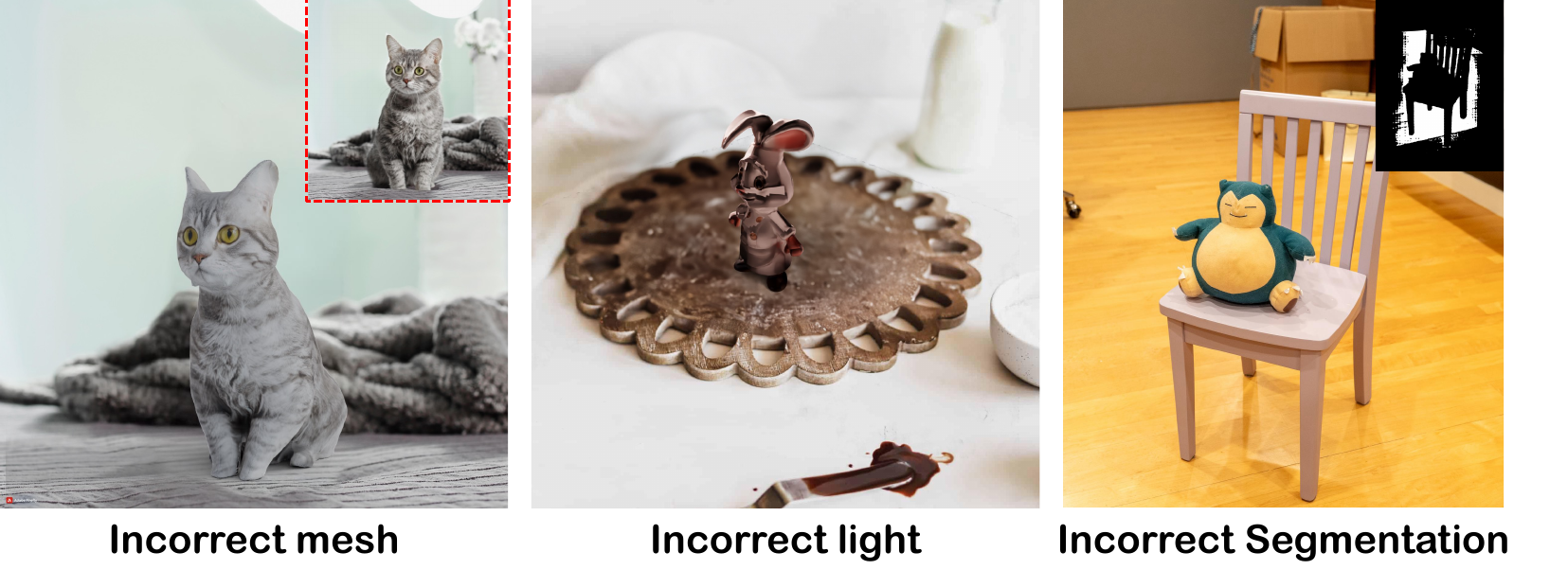}
\vspace{-0.8 cm}

\caption{\revision{\textbf{Representative failure cases of OMINI.}
The results may be degraded by inaccurate geometry from image-to-3D
reconstruction, incorrect environmental-illumination estimates, or imperfect
foreground segmentation.}}

\vspace{-0.7 cm}
\label{fig:failure_cases}
\end{figure}


\begin{table}[t]
\centering
\setlength{\tabcolsep}{5pt}
\renewcommand{\arraystretch}{1.1}

\caption{\textbf{\aqerevision{Progressive comparison of IllumiCorrector configurations.}}
CLIP-Image measures the appearance similarity between the input object
and its aligned reference-view rendering. Foreground Color Error (FCE) is the
mean CIEDE2000 ($\Delta E_{00}$) difference between their object regions.}
\vspace{-0.3cm}
\small
\begin{tabular}{@{}lcc@{}}
\toprule
\textbf{Configuration}
& CLIP-Image $\uparrow$
& \shortstack{FCE} $\downarrow$ \\
\midrule

DiffusionLight only
& \underline{0.9309}
& 21.78 \\

\aqerevision{Intensity optimization only}
& 0.9300
& \underline{12.82} \\

\textbf{Full IllumiCorrector}
& \textbf{0.9458}
& \textbf{7.73} \\

\bottomrule
\end{tabular}

\vspace{-0.6 cm}

\label{tab:ablation_illumicorrector}
\end{table}
\par\noindent\textbf{Parameter Sensitivity of IllumiCorrector.}
\revision{We examine the sensitivity of the HSL-based color correction in
IllumiCorrector to the percentile levels used to compute $T_S$ and $T_L$ and the interpolation weights $\lambda_s$ and $\lambda_l$. Each setting is varied individually while the others remain fixed. By default, $T_S$ and $T_L$ are computed at the 10th and 90th percentiles, respectively, while $\lambda_s=0.5$ and $\lambda_l=0.3$. The same defaults are used for all cases without case-specific tuning. To quantify the effect of each
parameter, we compute the percentage change in FCE relative to the default setting for each case and average it over cases. As shown in Fig.~\ref{fig:parameter_sensitivity}, the magnitude of the mean change remains within $1.6\%$ across all evaluated settings. Negative values indicate a lower FCE than the default setting. These results demonstrate that IllumiCorrector is stable across the tested parameter ranges.}

\par\noindent \textbf{Robustness of VLM-Guided Intensity Optimization.} We further investigate the stability of our VLM-guided intensity optimization by simulating diverse initialization scenarios, as visualized in Fig. \ref{fig:ablation_vlm}. Specifically, when initialized with different initial values of $\gamma$, the optimization trajectories exhibit a distinct convergence pattern. Unlike manual tuning or fixed-parameter methods that are sensitive to the starting state, our VLM-based agent dynamically assesses and corrects the exposure deviations at each step. For each iteration, the agent evaluates the rendered composition with respect to object exposure and contact-shadow visibility, and determines whether the illumination intensity should be increased, decreased, or retained. \aqerevision{The optimization terminates upon VLM acceptance or after at most 10 render--feedback evaluations.}

\subsection{Application} 
\label{sec:application}
\par Beyond geometric editing and animation, OMINI can render the refined 3D asset under corrected environmental illumination and composite the result into different background images, as shown in
Fig.~\ref{fig:application}. This capability suggests potential applications in virtual reality (VR) and augmented reality (AR), where inserted objects must remain visually coherent with their surrounding environments. By adapting the environmental illumination to each scene, OMINI produces compatible object shading and visually plausible shadows for both static and animated
results.

\section{Limitation and Future Work}
\label{sec:limitation}


\par Despite the success of our hybrid method, we acknowledge certain limitations primarily stemming from the underlying 3D generation capabilities.
OMINI relies on an external image-to-3D model for initialization, and therefore
remains affected by inaccurate geometry in cases with complex topology or
severe self-occlusion. \revision{Representative failure cases are shown in
Fig.~\ref{fig:failure_cases}. These upstream errors may therefore propagate to the final rendering.} \revision{The object-aware illumination correction may also
become less reliable for transparent or highly specular objects, whose observed
appearance does not provide stable color and lightness cues for environmental
illumination.} \aqerevision{TextuRefiner optimizes an appearance texture rather than
an illumination-disentangled albedo map. Shading retained in the
texture may therefore combine with the rendering illumination,
potentially causing inconsistent shading under strongly directional
lighting.}
\revision{Under the default $1024\times1024$ setting with 3,000
LoRA steps, the one-time asset construction takes approximately 45 minutes on
an NVIDIA RTX 3090 GPU.} Once
constructed, the refined asset can be reused for different manipulations with
only low-cost 3D operations and rendering.

\par Moving forward, we aim to expand the boundaries of this hybrid paradigm. Currently, our background handling remains static. We plan to integrate video generation models  to synthesize dynamic backgrounds, allowing for the reconstruction of full 3D scenes from single images to support complex camera movements. Our current formulation processes one target asset at a time, while the composition task creates multiple instances of the same refined asset. Extending OMINI to jointly manipulate distinct interacting objects requires further handling of object--object occlusion, collisions, and mutual shadowing, which we leave for future work.

\section{Conclusion}
\par In this work, we present OMINI, a unified framework for object
manipulation that supports both geometric editing and animation from a
single image. Unlike prior 3D-based methods that refine individual rendered
outputs in 2D, OMINI adopts a \textit{Refine-before-Render} design, in which the texture of the reconstructed 3D asset is refined before rendering. Both tasks are then performed by manipulating and rendering the same refined 3D asset, maintaining appearance consistency throughout the manipulation process. TextuRefiner refines rendered views under appearance and geometric guidance and uses them to optimize the texture map, recovering fine-grained texture details. IllumiCorrector combines object-aware HSL-based color correction with VLM-guided intensity optimization to produce visually plausible renderings with shadows. Extensive quantitative and qualitative experiments demonstrate that OMINI outperforms existing baselines on both geometric editing and animation tasks while supporting diverse object manipulation applications.

\section*{Acknowledgment} This work was supported by the Fundamental and Interdisciplinary Disciplines Breakthrough Plan of the Ministry of Education of China (No. JYB2025XDXM103).

\bibliographystyle{IEEEtran}
\bibliography{reference}

\begingroup
\makeatletter
\def\@IEEEBIOskipN{6pt}
\def\@textbottom{\vskip0pt plus 1fill}
\makeatother
\begin{IEEEbiography}[{\includegraphics[trim=180bp 20bp 180bp 110bp,clip,width=1in,height=1.25in,keepaspectratio]{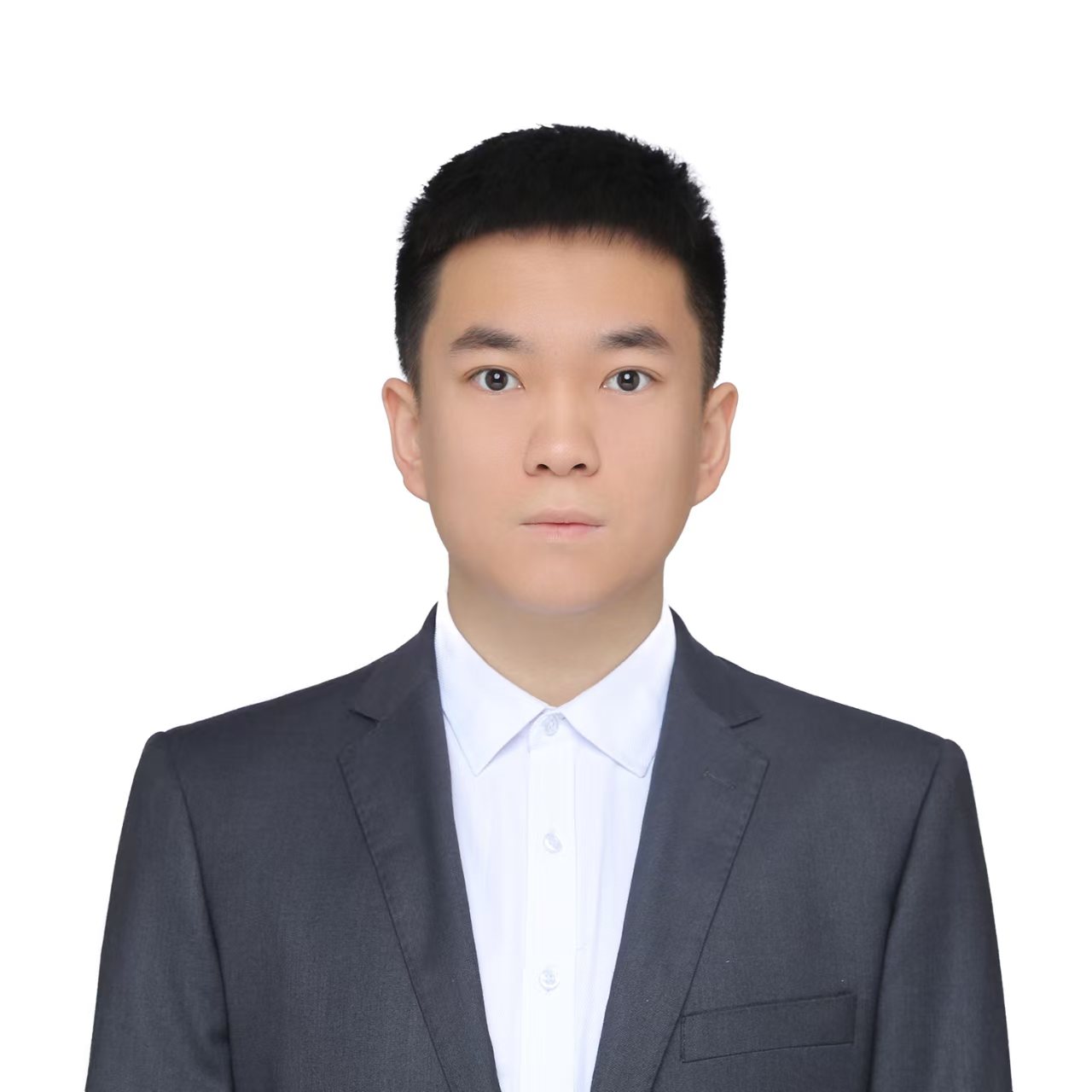}}]{Ruisi Zhao}
received the bachelor's degree from the University of Electronic Science and Technology of China (UESTC), Chengdu, China, in 2022. He is currently pursuing the Ph.D. degree in artificial intelligence with Zhejiang University, Hangzhou, China, under the supervision of Prof.~Yi~Yang. His research interests include computer vision and machine learning, especially generative models and 3D visual content creation.
\end{IEEEbiography}

\begin{IEEEbiography}[{\includegraphics[trim=160bp 16bp 160bp 65bp,clip,width=1in,height=1.25in,keepaspectratio]{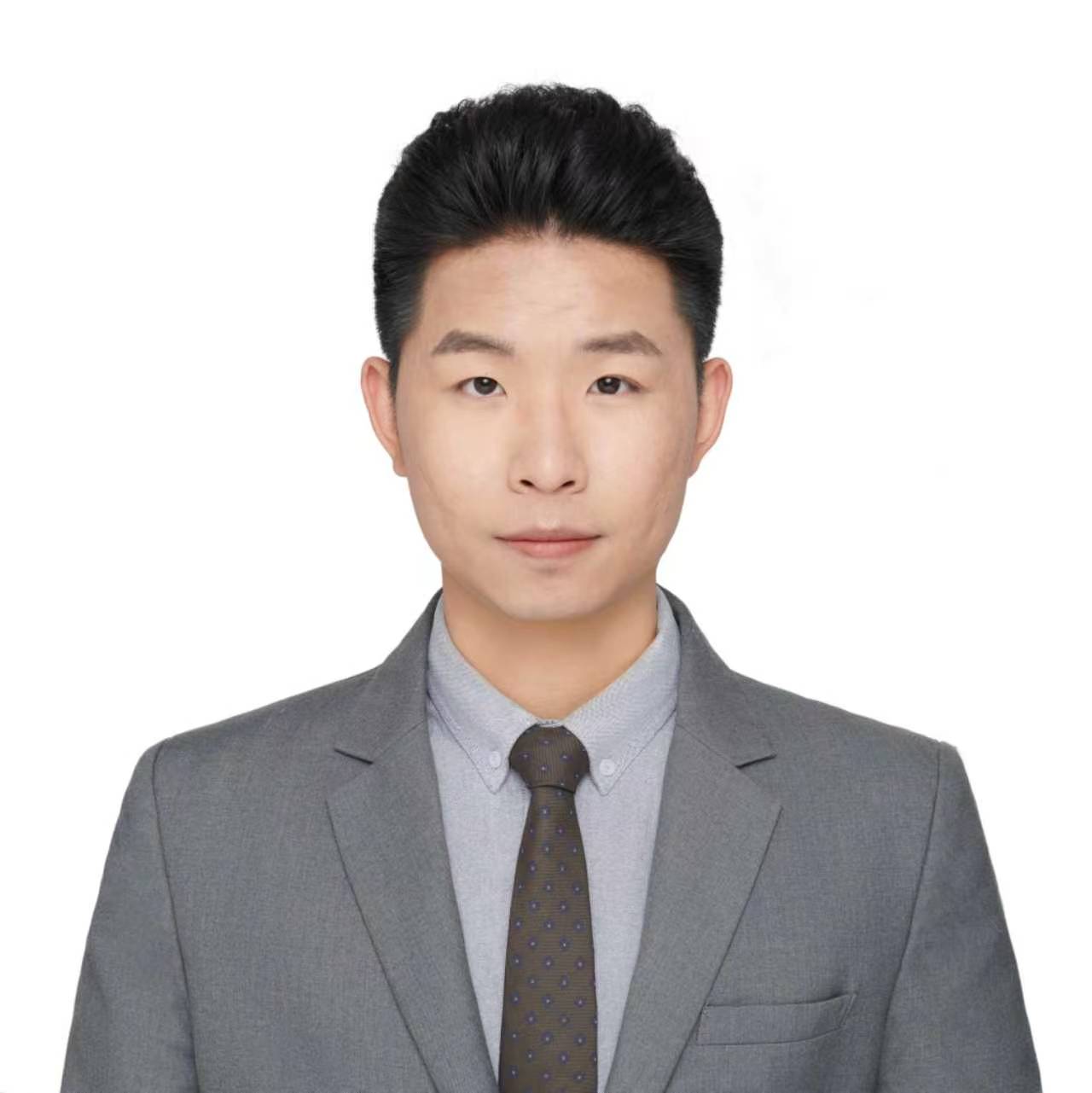}}]{Zechuan Zhang}
received the B.S. degree from Zhejiang University, Hangzhou, China, in 2023, where he is currently pursuing the Ph.D. degree. He has been a Visiting Research Scholar at the Broad Institute of MIT and Harvard, Cambridge, MA, USA. His research interests include computer vision, generative artificial intelligence, multimodal learning, and intelligent agents, with a focus on image and video generation, editing, and visual reasoning. He has published papers in leading international conferences and journals and served as a reviewer for conferences, including NeurIPS and AAAI.
\end{IEEEbiography}

\begin{IEEEbiography}[{\includegraphics[trim=36bp 0bp 36bp 90bp,clip,width=1in,height=1.25in,keepaspectratio]{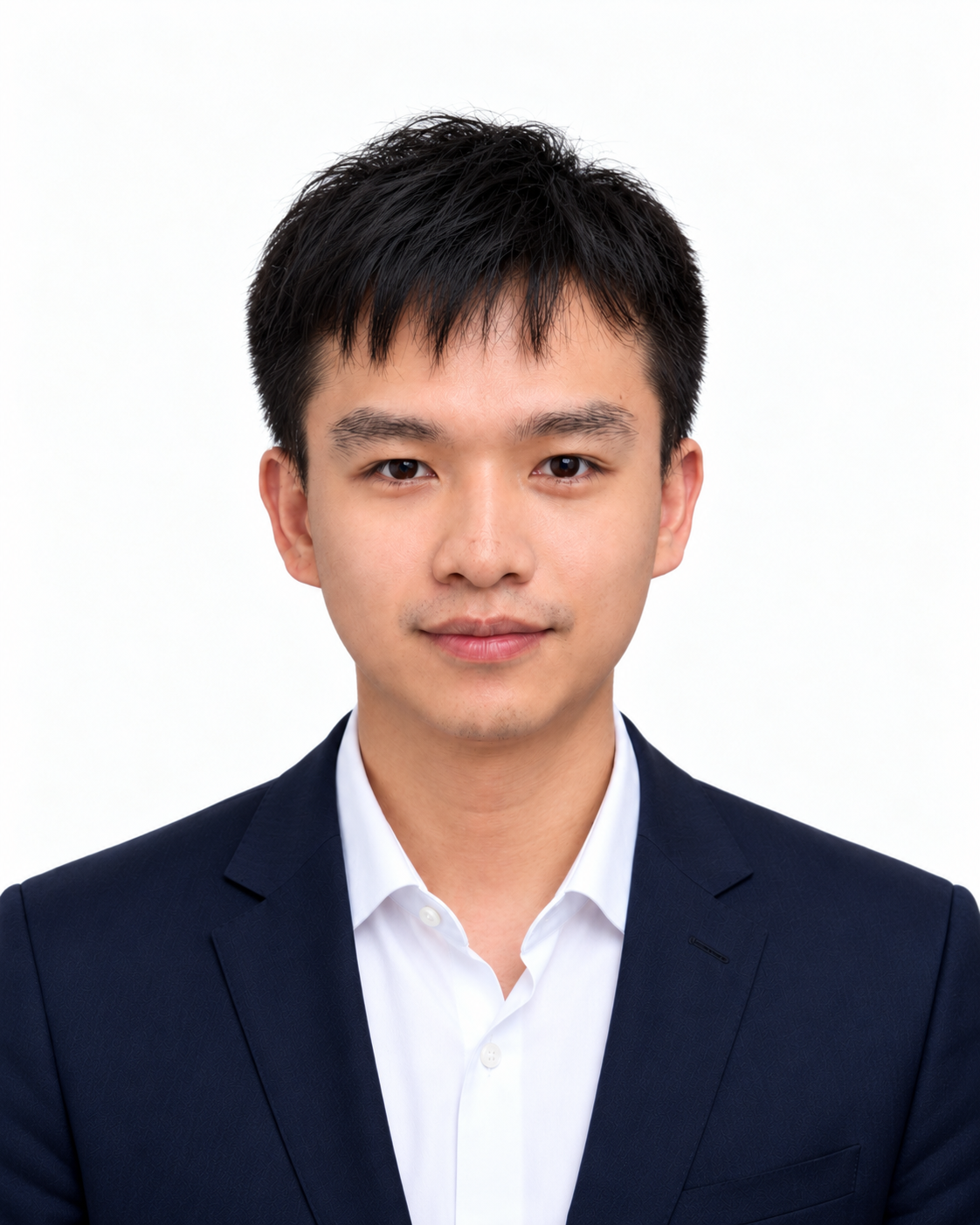}}]{Zongxin Yang}
received the B.Eng. degree from the University of Science and Technology of China, Hefei, China, in 2018, and the Ph.D. degree in computer science from the University of Technology Sydney, Sydney, Australia, in 2021. He is currently a ZJU100 Young Professor with the College of Artificial Intelligence, Zhejiang University, Hangzhou, China. Previously, he was a Research Fellow in the Department of Biomedical Informatics, Harvard Medical School, Harvard University, and a postdoctoral researcher at Zhejiang University. His research interests include multimodal understanding and generation, generative artificial intelligence, intelligent agents, and embodied intelligence.
\end{IEEEbiography}

\begin{IEEEbiography}[{\includegraphics[trim=24bp 0bp 24bp 70bp,clip,width=1in,height=1.25in,keepaspectratio]{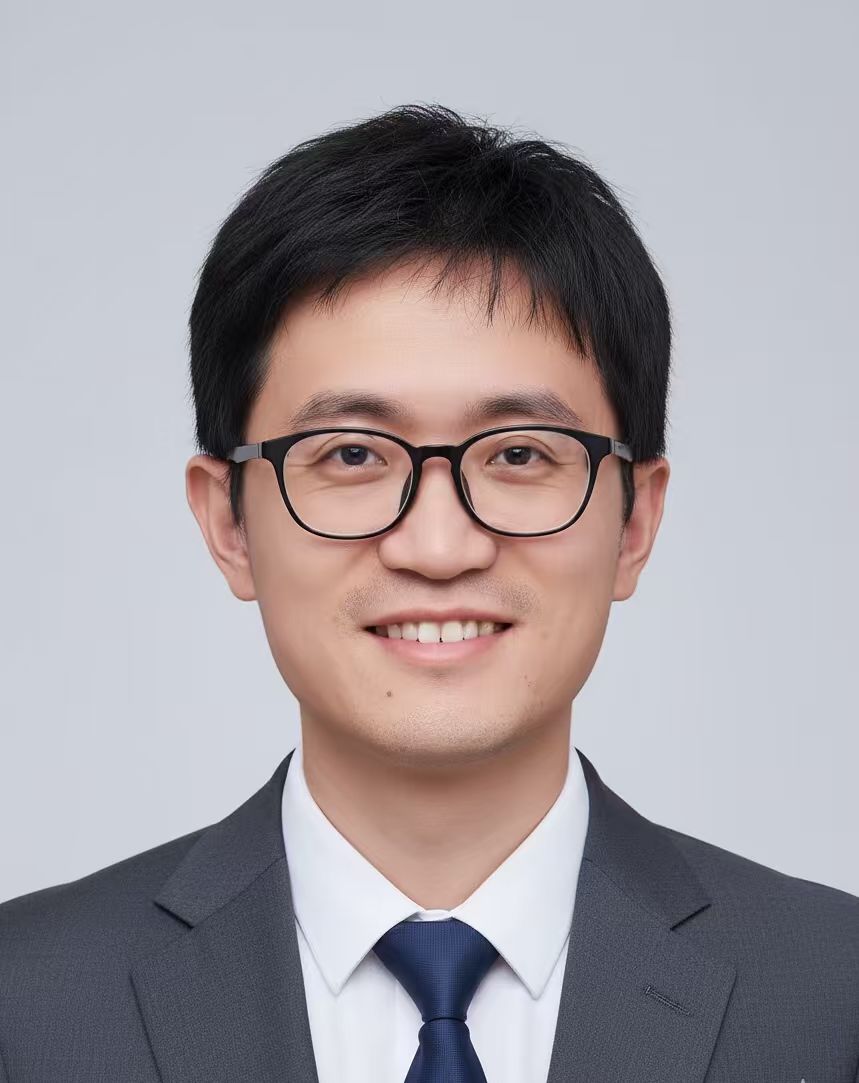}}]{Yi Yang}
(Fellow, IEEE) received the Ph.D. degree from Zhejiang University, Hangzhou, China, in 2010. He is currently a professor with Zhejiang University. He was a professor with the University of Technology Sydney. He was a postdoctoral researcher at the School of Computer Science at Carnegie Mellon University. His current research interests include machine learning and its applications to multimedia content analysis and computer vision, such as multimedia retrieval and video content understanding.
\end{IEEEbiography}
\par\vfill
\endgroup

\end{document}